\documentclass[10pt,twocolumn,letterpaper]{article}

\usepackage[pagenumbers]{cvpr} 

\definecolor{cvprblue}{rgb}{0.21,0.49,0.74}
\usepackage[pagebackref,breaklinks,colorlinks,allcolors=cvprblue]{hyperref}

\def\paperID{} 
\def\confName{CVPR}
\def\confYear{2026}

\title{E2E-GNet: An End-to-End Skeleton-based Geometric Deep Neural Network for Human Motion Recognition}

\author{
    Mubarak Olaoluwa \quad
    Hassen Drira \\[0.3em]
    University of Strasbourg, ICube Laboratory, UMR 7357, F-67412 Strasbourg, France \\
    {\tt\small \{molaoluwa, hdrira\}@unistra.fr} \; 
}

\usepackage{ulem}
\usepackage{makecell}
\usepackage{array,makecell}
\usepackage{booktabs} 
\usepackage{multirow}
\usepackage{amssymb}
\usepackage{bm}
\usepackage{amsmath}
\usepackage{algorithm, algpseudocode}
\usepackage{multirow}
\usepackage[table]{xcolor}
\usepackage{bbm}

\begin{document}
\maketitle
\begin{abstract}
Geometric deep learning has recently gained significant attention in the computer vision community for its ability to capture meaningful representations of data lying in a non-Euclidean space. To this end, we propose E2E-GNet, an end-to-end geometric deep neural network for skeleton-based human motion recognition. To enhance the discriminative power between different motions in the non-Euclidean space, E2E-GNet introduces a geometric transformation layer that jointly optimizes skeleton motion sequences on this space and applies a differentiable logarithm map activation to project them onto a linear space.  Building on this, we further design a distortion-aware optimization layer that limits skeleton shape distortions caused by this projection, enabling the network to retain discriminative geometric cues and achieve a higher motion recognition rate. We demonstrate the impact of each layer through ablation studies and extensive experiments across five datasets spanning three domains show that E2E-GNet outperforms other methods with lower cost. Code is available at \url{https://github.com/ayodejimb/E2E-GNet}.
\end{abstract}    
\section{Introduction}
\label{sec:intro}
Human motion recognition aims to automatically interpret and categorize human behaviors from visual observations, forming a core task in human-centered computer vision \cite{wang2018rgb, kale2016study}. Its importance spans a wide range of applications, including video surveillance, action recognition, human–robot collaboration, and analysis of musculoskeletal disorders in healthcare. Early studies approached this problem primarily through the analysis of still images or conventional RGB video sequences, relying heavily on 2D appearance cues such as color, texture, and silhouettes \cite{aggarwal1999human, turaga2008machine, zhu2016handcrafted}. However, these methods were inherently constrained by sensitivity to background clutter, occlusions, variations in illumination and viewpoint, and differences in movement execution—challenges that persisted even with the emergence of deep learning pipelines. \\
\indent Advances in affordable RGB-D sensors, such as the Microsoft Kinect \cite{zhang2012microsoft}, opened a new line of research by providing depth measurements that are less affected by lighting and offer richer 3D structural detail. This enabled more reliable estimation of body joint locations from depth maps, prompting the development of numerous RGB-D–based motion analysis frameworks \cite{aggarwal2014human, simonyan2014two, zhang2016rgb}. Building on these capabilities, skeleton-based motion recognition has attracted growing interest due to its compact, semantically meaningful representation of human movement. By modeling actions through 3D joint trajectories rather than raw pixel data, skeleton-based approaches mitigate background and appearance variability, provide viewpoint robustness, and focus directly on the geometric structure of human motion—making them particularly well-suited for modern learning-based recognition systems. \\
\indent The emergence of skeleton-based representations has motivated the development of geometric deep learning models capable of operating on the non-Euclidean structure inherent in the skeleton data. Skeleton-based manifold-aware deep architectures such as those leveraging Kendall’s shape space \cite{hosni2020geometric, friji2020geometric, zaier2025geometry, friji2023geometric, friji2021geometric}, Lie-group representations \cite{zaier2025geometry} or symmetric positive definite (SPD) matrices formulations \cite{nguyen2021geomnet} have demonstrated the value of incorporating geometric constraints into motion recognition pipelines. While these methods underscore the potential of geometry-driven learning, they also reveal two main limitations: (i) the lack of an end-to-end training pipeline that jointly optimizes both the geometric and deep learning components on the manifold, and (ii) distortion effects introduced during the projection of skeletons from their nonlinear manifold to linear (tangent) spaces. These issues hinder the expressive capacity and stability of geometric networks, especially when deployed across diverse domains. In this work, we address these issues and our contributions are as follows:
\begin{itemize}
    \item First, we propose E2E-GNet, an end-to-end geometric deep network with a new geometric transformation layer enabling manifold to Euclidean learning. 
    \item Second, we design a distortion-minimization layer that explicitly reduces the geometric distortions arising during manifold-to-tangent projections, thereby improving representation fidelity and model performance.
    \item Third, extensive experiments on five benchmarks across three domains---action recognition, disease analysis, and rehabilitation---show that E2E-GNet consistently outperforms other methods with lower computational overhead.
\end{itemize}
\section{Related Work}
\label{sec:stoa}

\subsection{Geometric Deep Learning}
\label{gdl_stoa}
Geometric deep learning has recently attained superior performance than conventional deep learning techniques in various domains such as 3D vision and graphics, medical imaging, and autonomous systems and robotics \cite{xiao2020survey}.  This is attributed to their capability to account for the underlying geometry of the input data which usually lies in a non-Euclidean space. Existing work on geometric deep learning can be categorized into two main research axes: techniques based on graphs and techniques based on manifolds \cite{cao2020comprehensive}. Many previous studies \cite{defferrard2016convolutional, atwood2016diffusion, verma2018feastnet, hanocka2019meshcnn, yan2017adaptive, li2018adaptive} have adapted convolution neural networks (CNN) to graphs for graph-based approaches. For brevity and since this study pertains to geometric deep learning techniques on manifolds, we focus on existing research in this area. We however refer readers who are interested in additional studies on graphs to these comprehensive surveys \cite{cao2020comprehensive, xiao2020survey, bronstein2017geometric, chen2024survey}. \\
\indent Existing work on manifold-based approaches can be categorically classified into two: feature space deep approaches \cite{chakraborty2020manifoldnet, masci2015geodesic, boscaini2015learning, boscaini2016learning, monti2017geometric, huang2018building, huang2017riemannian, huang2017deep, nguyen2019neural, nguyen2021geomnet} and shape space deep approaches \cite{friji2020geometric, hosni2020geometric, friji2021geometric, friji2023geometric, zaier2025geometry}. Since our method falls in the latter category, we briefly review this line of work. Friji et al. \cite{friji2021geometric, friji2020geometric} introduced deep models on Kendall shape space for human action recognition and later extended them to face and gait recognition in \cite{friji2023geometric}. Hosni et al \cite{hosni2020geometric} proposed a Kendall space autoencoder for gait analysis. Recently, Zaier et al. \cite{zaier2025geometry} proposed a transformer architecture on Kendall space and Lie group representations for human motion prediction. While these works have demonstrated the promise of deep learning on shape-space geometry, they share two key limitations: (i) end-to-end training of deep models on the manifold remains unresolved, and (ii) distortions introduced when projecting shapes to tangent space remain unaddressed. In this work, we offer an effective remedy to both, bridging these gaps in prior research. 

\subsection{Skeleton-based Human Motion Recognition}
\label{hmr_stoa}
\uline{\textbf{Action Recognition}:}
Skeleton-based human action recognition has evolved along three major methodological directions: GCN-based, transformer-based, and manifold-based methods. ST-GCN \cite{yan2018spatial} established the foundation of GCN-based by integrating graph convolutions for unified spatial–temporal modeling. Subsequent works largely follow this paradigm while strengthening the expressive power of GCNs \cite{shi2019two, liu2020disentangling, chen2021multi, chen2021channel, chi2022infogcn, duan2022pyskl, huang2023graph, zhou2023learning, xiang2023generative, lee2023hierarchically, xie2024dynamic, zhou2024adaptive}. Among recent advances, BlockGCN \cite{zhou2024blockgcn} introduced an efficient encoding scheme and graph convolution block to strengthen skeletal topology with reduced cost, and ProtoGCN \cite{liu2025revealing} proposed a prototype-learning strategy to capture subtle action variations. \\
\indent Prior transformer-based approaches leverage self-attention to capture long-range joint dependencies for action recognition \cite{plizzari2021skeleton, xin2023skeleton, zhang2021stst, do2024skateformer, pang2022igformer, gao2022focal}. Building on this direction, recent work \cite{zheng2024spatio} integrates GCNs with transformers to encode richer spatio-temporal representations, while \cite{ray2025autoregressive} combines hypergraph convolution with transformer blocks to aggregate complementary features for diverse actions. Furthermore, the study by \cite{wu2024frequency} introduces a frequency-based attention mechanism to improve the discrimination of subtle movements.\\
\indent Different to transformer-based methods, manifold-based approaches provide geometry-aware formulations for action recognition by representing skeletal motion in non-Euclidean spaces. Early work \cite{huang2017deep} leveraged Lie group representations to model skeletal dynamics on matrix manifolds, while subsequent efforts \cite{friji2021geometric} built deep architectures on Kendall’s shape space for learning pose actions. Additional studies \cite{nguyen2021geomnet} introduced neural networks on the manifold of symmetric positive definite matrices to capture discriminative action features. In this work, we introduce a Kendall shape space manifold-based deep approach that demonstrates consistent superior performance over prior manifold-based, transformer-based and GCN-based approaches. \\
\noindent \uline{\textbf{Disease and Rehabilitation Assessment}:}
In earlier times, studies \cite{lee2019learning, das2011quantitative, tao2016comparative, parmar2016measuring, elkholy2019efficient} on disease and rehabilitation assessments relied on extracting handcrafted features from input data to indicate abnormalities in motion or exercise. However, the arrival and impressive performance of various deep learning methods have replaced these traditional techniques with deep learning ones. Most recent studies have applied various deep learning methods for skeleton-based disease and exercise rehabilitation assessment on humans. Generally speaking, these works can be broadly put into two main categories: regression-based approaches \cite{williams2019assessment, liao2020deep, du2021assessing, mottaghi2022automatic, raihan2021automated, deb2022graph} and non-regression-based approaches \cite{bruce2020skeleton, bruce2021skeleton, yu2022egcn, bruce2024egcn++, kim2021patient, bai2025multiscale}. \\
\indent Since our work belongs to non-regression–based approaches, we briefly review existing methods in this category. Bruce et al. \cite{bruce2021skeleton} introduced the Elderly Home Exercise (EHE) dataset of Alzheimer’s patients and applied a GCN-based abnormality classifier built upon their earlier model \cite{ bruce2020skeleton}. Subsequent studies extended this line of work to ensemble GCN architectures for improved abnormality classification \cite{bruce2024egcn++, yu2022egcn}. More recently, Bai et al. \cite{bai2025multiscale} proposed an SPD-manifold method for abnormality assessments but at the cost of a higher model complexity. Compared to these approaches, our proposed method achieves superior accuracy while maintaining lower computational cost, highlighting a more efficient and effective design within non-regression frameworks. 

\section{Proposed End-to-End Geometric Network}
\label{e2e_framework}
\begin{figure*}[tp]
    \centering
    \includegraphics[width=\linewidth]{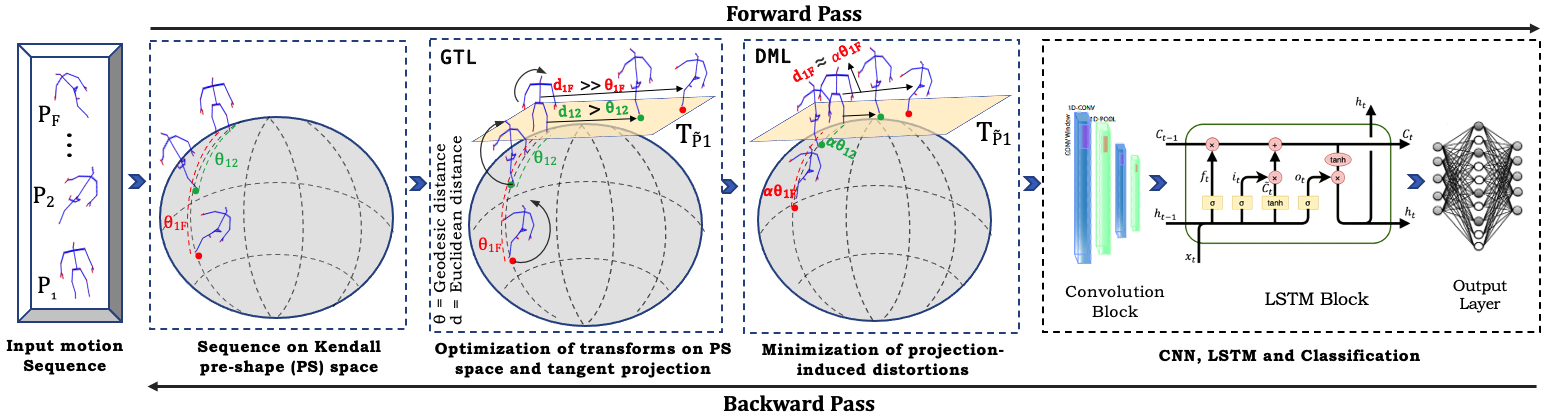}
    \caption{\textbf{Illustration of proposed E2E-GNet}. The input motion sequences are modeled onto the pre-shape space of a unit sphere. Then the geometric transformation layer (GTL) learns and applies optimal transforms to the sequences for improved feature extraction, with projection onto the tangent space via a log-map activation function. The distortion minimization layer (DML) then acts on these tangent representative skeletons to reduce projection-induced distortions. Finally, convolution and LSTM modules extract discriminative spatio-temporal features for classification, and the optimization is done in an end-to-end manner spanning both manifold and tangent spaces.}
    \label{main_figure}
\end{figure*}

Fig.~\ref{main_figure} presents an overview of the proposed E2E-GNet, and the following sections describe each component in detail.

\subsection{Modeling of Sequences on Pre-shape Space}
\label{modeling_method}
We use the position features of the skeleton motion sequences only. We begin by modeling these features onto a manifold space that is invariant to translation and scaling (also known as Kendall's pre-shape space \cite{kendall1984shape}). Each skeleton $X$ is denoted as $n$ joints in $\mathbbm{R}^3$, i.e., $X \in \mathbbm{R}^{n \times 3}$. Translation variability is removed by premultiplying $X$ by $H$ where $H$ denotes $(n-1) \times n$ submatrix of a Helmert matrix \cite{dryden2016statistical} but with the first row removed. Thus $HX \in \mathbbm{R}^{(n-1) \times 3}$. Scaling variability is removed from the centered skeleton $HX$ using the inner product and all skeletons were scaled to have a unit norm. Thus, we arrive at the pre-shape space $C \boldsymbol{=} \{ HX \in C_s |\, \|HX\|^2 \boldsymbol{=} (HX)^T (HX) \boldsymbol{=} 1 \}$ where $C_s$ is a centered and scaled skeleton. $C$ then becomes a $(3n-4)$-dimensional unit sphere as a result of the unit norm constraint.

\subsection{Geometric Transformation Layer (GTL)}
\label{gtl_method}
The skeleton motion sequences on the pre-shape space are received by the proposed geometric transformation layer (GTL) that passes these sequences from the non-linear pre-shape space to the tangent space. It does this in two steps: (i) Optimization over $SO(3)$ matrices applied to motion sequences and (ii) Projection of transformed sequences onto the tangent space through log map activation.  \\
\textit{\uline{Optimization over $SO(3)$ matrices}}: Let $P \in \mathbbm{R}^{\scriptsize F \times n \times 3}$ denote a sequence from pre-shape space $C$ with $\scriptsize F$ and $n$ representing sequences and joints respectively. For each frame $f \in \{1,\dots,F\}$, let $P_f \in \mathbbm{R}^{n \times 3}$ denote the $f$-th skeleton. We learn rotation parameters $\theta_f$ for each skeleton to generate a rotation matrix $R_f(\theta_f) \in SO(3)$. Each skeleton shape is then transformed as:
\begin{equation}
\widetilde{P}_f \boldsymbol{=} P_f\, R_f(\theta_f), \qquad R_f(\theta_f) \in SO(3).
\label{eq:rigid_constrained}
\end{equation}
Thus, we arrive at Kendall's shape space (space obtained by quotienting out rotational variability $C/SO(3)$) on the unit sphere. Each skeleton shape $\widetilde{P}_f$ belongs to a point in Kendall, and the distance between any two points indicates the magnitude of the difference between these shapes. We refer to this variant as \emph{Rigid-Constrained}. 

\noindent\uline{Logarithm map activation function}: We then introduce a differentiable Riemannian logarithm map as a non-linear geometric activation function to project the transformed skeleton sequences from the shape space to tangent space. 

Let $\widetilde{P}_f \in \mathbbm{R}^{n \times 3}$ denote the transformed skeleton at frame $f$, we project each skeleton onto the tangent space of the first skeleton frame $\widetilde{P}_{1}$, which is chosen as the reference configuration. The logarithm map activation between $\widetilde{P}_{f}$ and $\widetilde{P}_{1}$ is therefore derived as:  
\begin{equation}
\log_{\widetilde{P}_1}^{\sigma}(\widetilde{P}_f)
\boldsymbol{=} \widetilde{Z}_{f} \boldsymbol{=}
\frac{\theta_d}{\sin(\theta_d)} 
\left( 
\widetilde{P}_{f} - \cos(\theta_d)\,\widetilde{P}_{1} 
\right),
\quad \theta_d \neq 0.
\label{eq:logmap}
\end{equation}

where $\theta_d \boldsymbol{=} \arccos\!\left(\langle \widetilde{P}_{1},\, \widetilde{P}_{f} \rangle\right)$ is the geodesic distance between $\widetilde{P}_1$ and $\widetilde{P}_f$ and $\langle \cdot,\cdot \rangle$ denotes the Frobenius inner product. This operation maps all transformed skeletons to a common tangent space at $\widetilde{P}_{1}, (T_{\widetilde{P_1}})$, enabling a consistent linearized representation for subsequent learning.  

If one would like to go from the tangent space back to the shape space, one can use the Riemannian exponential map (Eq. \ref{eq:expmap}) and if one wishes to move between tangent spaces at different reference frames, one can apply parallel transport ~\cite{guigui2022numerical}. During the backward pass, we do not backpropagate through the exponential map since it is not part of the computational graph during the forward pass. 

\begin{multline}
\exp_{\widetilde{P}_1}(\widetilde{Z}_f)
= \widetilde{P}_f
= \cos(\|\widetilde{Z}_f\|_F)\,\widetilde{P}_1 \\
+ \frac{\sin(\|\widetilde{Z}_f\|_F)}{\|\widetilde{Z}_f\|_F}\,
\widetilde{Z}_f,
\quad \|\widetilde{Z}_f\|_F \neq 0.
\label{eq:expmap}
\end{multline}
where $\|\cdot\|_F$ denotes the Frobenius norm and $\widetilde{Z}_{f}$ is the representative skeleton on the tangent space.  

\subsection{Distortion Minimization Layer (DML)}
\label{dml_method}
Let us recall the logarithm map activation in Eq.~(\ref{eq:logmap}). The corrective factor $\bigl(\theta_d / \sin(\theta_d)\bigr)$ is not constant, and its variability introduces two forms of distortion. First, \textbf{global distortion}: when a skeleton $\widetilde{P}_f$ lies at a non-zero distance from the reference skeleton $\widetilde{P}_{1}$, the geodesic distance $\theta_d$ is $> 0$ and increases with the separation between the two skeletons. Since $\sin(\theta_d) \leq \theta_d \; \forall \; \theta_d > 0$, the ratio $(\theta_d / \sin(\theta_d))$ exceeds~1 and grows with~$\theta_d$. As a result, the magnitude of the representative skeleton $\widetilde{Z}_f$, that is $\bigl\| \widetilde{Z}_{f}\bigr\|$ on the tangent space is stretched, meaning that $\bigl\| \widetilde{Z}_{f}\bigr\|$ in the tangent space overestimate the true geodesic distance~$\theta_d$ on the manifold and which becomes worse for larger $\theta_d$. Second, \textbf{pairwise distortion}: this global stretching relative to $\widetilde{P}_{1}$ propagates to the relationships among all other projected skeletons. For two skeletons $\widetilde{P}_{2}$ and $\widetilde{P}_{3}$, the distance between their tangent representations at $\widetilde{P}_{1}, \bigl\|\ \widetilde{Z}_{2}-\widetilde{Z}_{3}\bigr\|$ generally differs from their true geodesic distance on the manifold. That is, $
\bigl\| \widetilde{Z}_{2} - \widetilde{Z}_{3} \bigr\| \neq \theta_{d}(\widetilde{P}_{2}, \widetilde{P}_{3})$. Hence, both absolute distances to the reference and pairwise relationships among projected skeletons become distorted in the tangent space. \\
\indent To minimize these distortions, we introduce a learnable positive parameter $\alpha$, initialized in the interval $(-1,1)$ and constrained to remain positive during training through Euler's exponentiation. Each tangent representative skeleton $\widetilde{Z}_{f}$ is uniformly scaled such that:
\begin{equation}
\widetilde{Z}_{f} \;\mapsto\; \alpha \widetilde{Z}_{f}  
\quad \alpha > 0, \quad f \in \{1,\dots,F\}   
\end{equation}

\noindent thereby modulating the magnitude of $\widetilde{Z}_{f}$. Since $\widetilde{Z}_{f}$ is a horizontal tangent vector on Kendall shape space (i.e., orthogonal to rotational directions in the preshape space), multiplying it by the scalar $\alpha$ maintains orthogonality to rotation directions and leaves its direction $\widetilde{Z}_{f} / \|\widetilde{Z}_{f}\|_F$ unchanged. The geometric interpretation follows directly from the exponential map. Let $\widetilde{Z}_{f} \boldsymbol{=} \log_{\widetilde{P}_1}^{\sigma}(\widetilde{P}_f)$ and define the geodesic:
\[
\gamma(t) \boldsymbol{=} \text{exp}_{\widetilde{P}_{1}}(t\widetilde{Z}_{f})
\]
with $\gamma(0)=\widetilde{P}_{1}$, $\dot{\gamma}(0)\boldsymbol{=}\widetilde{Z}_{f}$, and $\gamma(1)=\widetilde{P}_{f}$. Then
\[
\text{exp}_{\widetilde{P}_{1}}(\alpha \widetilde{Z}_{f}) \boldsymbol{=} \gamma(\alpha), 
\]
showing that scaling the tangent representation corresponds to evaluating the same geodesic at parameter $\alpha$. Thus, the geodesic path is preserved; only the traveled geodesic distance becomes $\alpha \theta_d$ because the intrinsic deformation direction remains the same: 
\[
\frac{\alpha \widetilde{Z}_{f}}{\|\alpha \widetilde{Z}_{f}\|}_F \boldsymbol{=} \frac{\widetilde{Z}_{f}}{\|\widetilde{Z}_{f}\|}_F
\]
Moreover, curvature in Kendall shape space is determined by its Riemannian metric inherited from the preshape space via the quotient by rotations. Reparameterizing the geodesic by the learnable $\alpha$ does not alter the metric or curvature; it only changes the shape position along the same curved trajectory. Hence, the intrinsic geometry and deformation mode are preserved.

Since $\alpha > 0$, it follows that the optimal $\alpha$ satisfies $0 < \alpha < 1$, thus the displacement from $\widetilde{P}_{1}$ satisfies $\|\alpha \widetilde{Z}_{f}\|\approx \alpha\theta_d$, keeping the skeleton shape closer to the reference where the logarithm map provides a better linear approximation of the manifold. Consequently, $\alpha$ acts as an adaptive geodesic distance contraction that preserves direction and intrinsic curvature while reducing distortion in the tangent representation. We refer to this variant as \emph{Global Homogeneous} since the same $\alpha$ is applied uniformly to all joints and frames of the skeletal configuration.

\subsection{Additional GTL and DML Variants}
\label{gtl_trans_others}
To allow more flexibility for the deep architecture, we extend beyond rigid-constrained transformations and global homogeneous scaling, and explore alternative variants that relax strict geometric constraints, described as follows: 

\noindent \textbf{(a) GTL Variants}

\noindent \textit{\uline{Rigid-Unconstrained:}} In this setting, we directly learn a transformation matrix $A_f \in \mathbbm{R}^{3 \times 3}$ for each skeleton. Unlike the rigid–constrained case, $A_f$ is \emph{not} constrained to satisfy orthogonality during optimization. Each skeleton is transformed as $\widetilde{P}_f \boldsymbol{=} P_f\, A_f, \quad A_f \in \mathbbm{R}^{3 \times 3} $

\noindent \textit{\uline{Non-Rigid–Constrained:}} For this variant, we learn rotation parameters $\theta_{f,i}$ for each joint $i \in \{1,\dots,n\}$ in frame $f$ to generate joint-wise rotation matrices $R_{f,i}(\theta_{f,i}) \in SO(3)$ such that each joint is rotated independently. The transformed skeleton is given as $\widetilde{P}_{f,i} \boldsymbol{=} P_{f,i}\, R_{f,i}, \quad R_{f,i} \in SO(3),\ i=1,\dots,n. $

\noindent \textit{\uline{Non-Rigid–Unconstrained:}} In this setting, we learn a transformation matrix $A_{f,i} \in \mathbbm{R}^{3 \times 3}$ for each joint $i \in \{1,\dots,n\}$ in frame $f$. Unlike the constrained variant, each $A_{f,i}$ is \emph{not} required to satisfy orthogonality. Each joint is transformed independently as $\widetilde{P}_{f,i} \boldsymbol{=} P_{f,i}\, A_{f,i}, \quad A_{f,i} \in \mathbbm{R}^{3 \times 3},\ i=1,\dots,n. $

\noindent \textbf{(b) DML Variants}

\noindent \textit{\uline{Global inhomogeneous:}} For this variant, we learn $\alpha_i$ for each joint $i \in \{1,\dots,n\}$ and the same $\alpha_i$ is shared across all skeleton frames. 

\noindent \textit{\uline{Local homogeneous:}} In this variant, we learn $\alpha_f$ for each skeleton frame $f \in \{1,\dots,F\}$ and the same $\alpha_f$ is shared across all joints within that frame.

\noindent \textit{\uline{Local inhomogeneous:}} In this setting, we learn $\alpha_{f,i}$ for each joint $i \in \{1,\dots,n\}$ in each skeleton frame $f \in \{1,\dots,F\}$. 

\subsection{Feature Extraction and Classification}
\label{fea_and_classif_method}
For feature extraction, we use two Conv1D layers followed by a MaxPool1D and an LSTM, and for classification we apply two fully connected layers (FCL). We use two Conv1D layers for action datasets and one Conv1D layer for others. For all datasets, the Conv1D kernel size is set to 3 and the LSTM comprises 12 units. The hidden dimension of FCL is set to 512 for action datasets and 32 for others. 
\section{Experimental Datasets and Settings}
\label{datasets_and_settings}
\uline{\textbf{Datasets:}} We validate E2E-GNet on five diverse datasets described below: \\
\textbf{NTU RGB+D (NTU-60)} dataset \cite{shahroudy2016ntu} is a large-scale benchmark for human action recognition, comprising 56,880 skeleton-based action sequences performed by 40 different subjects and categorized into 60 action classes. Each sequence is represented by 25 body joints and includes at most two subjects. The dataset provides two standard evaluation protocols: (1) Cross-Subject (X-Sub) — training samples are collected from 20 subjects, while the remaining 20 are used for testing; and (2) Cross-View (X-View) — training samples are captured from camera views 2 and 3, with testing samples from camera view 1. \\
\textbf{NTU RGB+D 120 (NTU-120)} dataset \cite{liu2019ntu} is an extension of NTU RGB+D, containing 114,480 skeleton-based action sequences across 120 action classes. These actions are performed by 106 subjects and recorded using 32 distinct camera setups. The dataset also includes two standard evaluation protocols: (1) Cross-Subject (X-Sub) — training data are taken from 53 subjects, while the remaining 53 subjects are used for testing; and (2) Cross-Setup (X-Set) — training data are collected from 16 even-numbered setup IDs, and testing data from 16 odd-numbered setup IDs. \\
\textbf{Elderly Home Exercise (EHE)} dataset \cite{bruce2021skeleton} is a real-world Alzheimer's disease (AD) dataset collected in an elderly home. The dataset contains skeleton features of 25 subjects performing six daily morning exercises in a repetitive manner. There are 869 exercise sequences in total and 10 of the 25 subjects have been diagnosed with AD. \\ 
\textbf{Kinematic Assessment of Movement for Remote Monitoring of Physical Rehabilitation (KIMORE)} dataset \cite{capecci2019kimore} is another real-world dataset that includes exercise data from 78 subjects across three groups: a control group of experts (CG-E, 12 physiotherapists and experts in the rehabilitation of back pain and postural disorders), a control group of non-experts (CG-NE, 32 subjects), and a group with pain and postural disorders (GPP, 34 subjects with conditions such as Parkinson’s, stroke, and back pain). Participants performed five exercises and we use the skeleton features from \cite{yu2022egcn}, comprising 1,769 exercise sequences. \\
\textbf{The University of Idaho Physical Rehabilitation Movements (UI-PRMD)} dataset \cite{vakanski2018data} is a rehabilitation dataset that contains skeleton features from 10 healthy subjects with each subject performing 10 repetitions of 10 rehabilitation exercises. Each subject performed each exercise in both correct and incorrect manners. We use the version from \cite{yu2022egcn} where samples due to measurement errors and exercise performance with wrong limbs have been removed. This version contains 1,326 exercise sequences in total. 

\noindent \uline{\textbf{Pre-processing:}} For action datasets, we follow the data pre-processing steps in \cite{friji2021geometric} where we applied cubic spline interpolation independently to each subject in train and test to resample skeleton trajectories into smooth, equally-spaced sequences of fixed length. For other datasets (EHE, KIMORE and UI-PRMD), all train and test subjects in each dataset contain the same sequence length. \\
\noindent \uline{\textbf{Evaluation:}} For action datasets, we follow the same standard train–test split protocol used by all compared methods. For others (EHE, KIMORE and UI-PRMD), we use the same cross-subject five-fold cross-validation division protocol as done in all compared methods (see Appendix C of supplementary text). \\ 
\noindent \uline{\textbf{Settings:}} All experiments are conducted with PyTorch using Adam optimizer, fixed learning rate, cross-entropy loss and on a workstation with 64GB RAM and Tesla V100 GPU. Other settings and sequence length are given in Table \ref{other_exp_settings}. 
\begin{table}[ht]
\vspace{-0.3cm}
\centering
\setlength{\tabcolsep}{3.5pt}
\caption{Datasets other experimental settings}
{\footnotesize
\begin{tabular}{lcccc}
\hline
\textbf{Parameters} & \textbf{NTU-60/120} & \textbf{EHE} & \textbf{KIMORE} & \textbf{UI-PRMD} \\
\hline
Batch size         & 64                 & 12            & 16               & 16 \\
Epochs             & 50                 & 35            & 50               & 40 \\
Learning rate      & 1e-4               & 1e-3          & 1e-3             & 1e-3 \\
Sequence Length      & 100              & 221          & 150             & 150 \\
\hline
\end{tabular}
}
\label{other_exp_settings}
\end{table}

\section{Results and Discussion}
\label{results_and_discussion}
\subsection{Comparison with State-of-the-Art Methods}
\label{SOTA_comparison}

\definecolor{softblue}{RGB}{180,225,255}
\begin{table*}[t]
\centering
\setlength{\tabcolsep}{6.5pt}
\caption{Performance comparison (\% classification accuracy and cost) against SOTA methods on NTU-60 and NTU-120 datasets.}
{\footnotesize
\begin{tabular}{lcccccccc}
\hline
\multirow{2}{*}{\textbf{Category}} & 
\multirow{2}{*}{\textbf{Methods}} & 
\multirow{2}{*}{\textbf{Venue}} & 
\multirow{2}{*}{\textbf{Params (M)}} & 
\multirow{2}{*}{\textbf{FLOPs (G)}} & 
\multicolumn{2}{c}{\textbf{NTU-60 (\%)}} & 
\multicolumn{2}{c}{\textbf{NTU-120 (\%)}} \\
\cline{6-9}
 &  &  &  &  & \textbf{X-Sub} & \textbf{X-View} & \textbf{X-Sub} & \textbf{X-Setup} \\
\hline
\multirow{14}{*}{\makecell[c]{Graph\\Convolution}} 
 & ST-GCN \cite{yan2018spatial} & AAAI 2018 & 3.08 & 5.34 & 81.5 & 88.3 & 70.7 & 73.2 \\
 & 2s-AGCN \cite{shi2019two} & CVPR 2019 & – & – & 88.5 & 95.1 & 82.5 & 84.2 \\
 & MS-G3D \cite{liu2020disentangling} & CVPR 2020 & 3.17 & 10.27 & 91.5 & 96.2 & 86.9 & 88.4 \\
 & MST-GCN \cite{chen2021multi} & AAAI 2021 & 3.10 & – & 91.5 & 96.6 & 87.5 & 88.8 \\
 & CTR-GCN \cite{chen2021channel} & ICCV 2021 & 1.46 & 1.97 & 92.4 & 96.8 & 88.9 & 90.6 \\
 & InfoGCN \cite{chi2022infogcn} & CVPR 2022 & 1.60 & 1.84 & 93.0 & 97.1 & 89.8 & 91.2 \\
 & PYSKL \cite{duan2022pyskl} & ACM MM 2022 & 1.39 & 2.80 & 92.6 & 97.4 & 88.6 & 90.8 \\
 & SkeletonGCL \cite{huang2023graph} & ICLR 2023 & – & – & 92.8 & 97.1 & 89.8 & 91.2 \\
 & FR Head \cite{zhou2023learning} & CVPR 2023 & 1.99 & – & 92.8 & 96.8 & 89.5 & 90.9 \\
 & GAP \cite{xiang2023generative} & ICCV 2023 & – & – & 92.9 & 97.0 & 89.9 & 91.1 \\
 & HD-GCN \cite{lee2023hierarchically} & ICCV 2023 & 1.68 & 1.60 & 93.4 & 97.2 & 90.1 & 91.6 \\
 & DS-GCN \cite{xie2024dynamic} & AAAI 2024 & – & – & 93.1 & 97.5 & 89.2 & 91.1 \\
 & BlockGCN \cite{zhou2024blockgcn} & CVPR 2024 & 1.30 & 1.63 & 93.1 & 97.0 & 90.3 & 91.5 \\
 & ProtoGCN \cite{liu2025revealing} & CVPR 2025 & – & – & 93.8 & 97.8 & 90.9 & 92.2 \\
\hline
\multirow{5}{*}{\makecell[c]{Transformer/\\Graph Transformer}} 
 & IGFormer \cite{pang2022igformer} & ECCV 2022 & – & – & 93.4 & 96.5 & 85.4 & 86.5 \\
 & FG-STFormer \cite{gao2022focal} & ACCV 2022 & – & – & 92.6 & 96.7 & 89.0 & 90.6 \\
 & MAMS \cite{mao2023masked} & ICCV 2023 & – & – & 93.1 & 97.5 & 90.0 & 91.3 \\
 & SkateFormer \cite{do2024skateformer} & ECCV 2024 & 2.03 & 3.62 & 93.5 & 97.8 & 89.8 & 91.4 \\
 & JT-GraphFormer \cite{zheng2024spatio} & AAAI 2024 & – & – & 93.4 & 97.5 & 89.9 & 91.7 \\
 & FreqMixFormer \cite{wu2024frequency} & ACM MM 2024 & 2.04 & 2.40 & 93.6 & 97.4 & 90.5 & 91.9 \\
\hline
\multirow{3}{*}{\makecell[c]{HyperGraph\\Convolution}}
& DC-GCN+ADG \cite{cheng2020decoupling} & ECCV 2020 & – & 16.20 & 90.8 & 96.6 & 86.5 & 88.1 \\
& DST-HCN \cite{wang2023dynamic} & ICME 2023 & 2.93 & 3.50 & 92.3 & 96.8 & 88.8 & 90.7 \\
& AdaptiveHGCN \cite{zhou2024adaptive} & ICCV 2025 & 1.10 & 1.63 & 93.3 & 97.4 & 90.5 & 91.7 \\
\hline
\multirow{2}{*}{\makecell[c]{HyperGraph\\Transformer}}
 & Hyperformer \cite{zhou2022hypergraph} & ArXiv 2022 & 2.6 & 14.8 & 92.9 & 96.5 & 89.9 & 91.3 \\
 & AutoregAd-HGformer \cite{ray2025autoregressive} & WACV 2025 & 3.20 & 15.4 & 94.2 & 97.8 & 91.0 & 92.4 \\
\hline
\multirow{4}{*}{\makecell[c]{Deep\\Riemannian}}
 & LieNet \cite{huang2017deep} & CVPR 2017 & – & – & 61.4 & 67.0 & – & – \\
 & KShapeNet \cite{friji2021geometric} & ICCV 2021 & 0.93 & 0.01 & 97.0 & 98.5 & 90.6 & 86.7 \\
 & GeomNet \cite{nguyen2021geomnet} & ICCV 2021 & – & – & 93.6 & 96.3 & 86.5 & 87.6 \\
& \cellcolor{softblue}\textbf{E2E-GNet (Ours)} & \cellcolor{softblue}
& \cellcolor{softblue} \textbf{0.93}
& \cellcolor{softblue} \textbf{0.01}
& \cellcolor{softblue}\textbf{97.1} 
& \cellcolor{softblue}\textbf{98.6} 
& \cellcolor{softblue}\textbf{95.2} 
& \cellcolor{softblue}\textbf{93.3} \\
\hline
\end{tabular}
}
\label{ntu_SOTA_comparison}
\end{table*}

\begin{table*}[t]
\centering
\setlength{\tabcolsep}{15pt}
\caption{Performance comparison (\% classification accuracy and cost) against SOTA methods on EHE, KIMORE and UI-PRMD datasets.}
{\footnotesize
\begin{tabular}{lcccccc}
\hline
\textbf{Methods} & \textbf{Venue} & \textbf{Params (M)} & \textbf{FLOPs (G)} & \textbf{EHE} & \textbf{KIMORE} & \textbf{UI-PRMD}\\
\hline
SPDNet \cite{huang2017riemannian} & AAAI 2017 & – & – & 78.76 & 82.57 & 88.87 \\
GCN \cite{bruce2020skeleton} & TransAI 2020 & 3.10 & 3.80 & 77.63 & 76.27 & 75.44 \\
KShapeNet \cite{friji2021geometric} & ICCV 2021 & 0.07 & 0.003 & 83.26 & 76.60 & 88.22 \\
EGCN \cite{yu2022egcn} & IJCAI 2022 & 6.20 & 7.60 & 78.87 & 80.14 & 86.90 \\
EGCN++ \cite{bruce2024egcn++} & TPAMI 2024 & 6.20 & 7.60 & 86.38 & 83.56 & 89.95 \\
ManSPD \cite{bai2025multiscale} & SciReports 2025 & 267.20 & 0.60 & 84.74 & 85.05 & 92.40 \\
\cellcolor{softblue}\textbf{E2E-GNet (Ours)} & \cellcolor{softblue} & \cellcolor{softblue}\textbf{0.07} & \cellcolor{softblue}\textbf{0.003} & \cellcolor{softblue}\textbf{87.14} & \cellcolor{softblue}\textbf{85.93} & \cellcolor{softblue}\textbf{95.19} \\
\hline
\end{tabular}
}
\label{other_SOTA_comparison}
\end{table*}

Tables \ref{ntu_SOTA_comparison} and \ref{other_SOTA_comparison} present a comprehensive comparison of E2E-GNet against state-of-the-art (SOTA) methods across all benchmarks. On NTU-60, E2E-GNet outperforms SOTA performance \cite{friji2021geometric} by 0.1\% on both X-Sub and X-View protocols. Notably, on the more challenging NTU-120 dataset, E2E-GNet delivers substantial gains, outperforming SOTA \cite{ray2025autoregressive} by 4.2\% on X-Sub and 0.9\% on X-Setup. In the disease and rehabilitation domain, E2E-GNet again shows consistent superiority, surpassing SOTA \cite{bruce2024egcn++, bai2025multiscale} by 0.76\%, 0.88\%, and 2.79\% on EHE, KIMORE, and UI-PRMD, respectively. \\
\indent Beyond accuracy, E2E-GNet also demonstrates computational efficiency. Across all datasets, it maintains approximately the same computational cost with \cite{friji2021geometric}. This lower cost is further supported by lower inference time provided in Appendix A of the supplementary text. These results together showcase the effectiveness and computational efficiency of proposed E2E-GNet across diverse applications.

\subsection{Ablation Studies}
\label{ablation_study}
\textbf{\uline{Contribution of Components}:} We study the impact of proposed GTL and DML in E2E-GNet as shown in Tab.~\ref{contribution_ablation}. Baseline (BL) denotes Conv1D and LSTM directly on pre-shape space without GTL and DML. BL results show modest but less competitive performance across all datasets. \\ 
\indent Adding GTL to BL consistently boosts performance. For example, improvements of 6.7\% on NTU-120 X-Set, 5.44\% on KIMORE and 5.66\% on UI-PRMD. These gains result from enhanced discriminative feature extraction through optimal skeleton transformations and the convenient linear-space representation provided by the log-map activation.  \\ 
\indent Finally, adding DML to BL and GTL yields E2E-GNet which consistently improves performance across all datasets. For example, 4.9\%, 5.84\% and 8.13\% improvements on NTU-120 X-Set, KIMORE and UI-PRMD respectively. These results highlight the effectiveness of DML with a learning-based approach. Also, the end-to-end integration of GTL and DML significantly enhances the representation and discriminative power of the proposed E2E-GNet.

\begin{table}[ht]
\centering
\setlength{\tabcolsep}{1.2pt}
\caption{Ablation results (\% accuracy) on the contribution of each proposed layer in E2E-GNet. \textbf{E2E-GNet} is \textbf{BL+GTL+DML}.}
{\footnotesize
\begin{tabular}{lcccc|cc|c}
\toprule
\multirow{3}{*}{\textbf{Protocol}} 
& \multicolumn{4}{c|}{\textbf{Action}} 
& \multicolumn{2}{c|}{\textbf{Disease}} 
& \multicolumn{1}{c}{\textbf{RHB}} \\ 
\cmidrule(lr){2-5} \cmidrule(lr){6-7} \cmidrule(lr){8-8}
& \multicolumn{2}{c}{\textbf{NTU-60}} 
& \multicolumn{2}{c|}{\textbf{NTU-120}} 
& \textbf{EHE} & \textbf{KIMORE} & \textbf{UI-PRMD} \\
\cmidrule(lr){2-3} \cmidrule(lr){4-5}
& \textbf{\scriptsize{X-Sub}} & \textbf{\scriptsize{X-View}} 
& \textbf{\scriptsize{X-Sub}} & \textbf{\scriptsize{X-Set}} 
&  &  & \\ 
\midrule
\textbf{\scriptsize{Baseline (BL)}} & 93.2 & 95.8 & 89.1 & 81.7 & 81.69 & 74.65 & 81.40 \\
\textbf{\scriptsize{BL+GTL}} & 95.2 & 96.8 & 92.3 & 88.4 & 83.15 & 80.09 & 87.06 \\
\cellcolor{softblue} \textbf{\scriptsize{BL+GTL+DML}} & \cellcolor{softblue} \textbf{97.1} & \cellcolor{softblue} \textbf{98.6} & \cellcolor{softblue} \textbf{95.2} & \cellcolor{softblue} \textbf{93.3} & \cellcolor{softblue} \textbf{87.14} & \cellcolor{softblue} \textbf{85.93} & \cellcolor{softblue} \textbf{95.19} \\
\hline
\end{tabular}
}
\label{contribution_ablation}
\end{table}
\noindent \textbf{\uline{Performance Comparison of GTL variants}:} We study the impact of each transformation variant by adding it to the baseline only, reported in Tab.~\ref{transformation_matrices_results}. We remark a consistent variant for each domain. For example, on action datasets, the non-Rigid variant (unconstrained) yields the best performance, while for disease and rehabilitation datasets, the rigid variant proves the most optimal, particularly with $SO(3)$ constraints for disease datasets. This explains that action datasets contain highly variable, joint-specific articulations, so non-rigid transformations provide the flexibility needed to model these heterogeneous motions. In contrast, disease and rehabilitation movements are more globally coherent and biomechanically constrained, making rigid transformations the best variant for this application.
\begin{table}[ht]
\centering
\setlength{\tabcolsep}{1.7pt}
\caption{Performance (\% accuracy) of GTL variants on all datasets. Best are bolded. RHB is Rehabilitation. R, NR, Ctr and Utr are rigid, non-rigid, constrained and unconstrained. }
{\footnotesize
\begin{tabular}{lcccc|cc|c}
\toprule
\multirow{3}{*}{\textbf{Variants}} 
& \multicolumn{4}{c|}{\textbf{Action}} 
& \multicolumn{2}{c|}{\textbf{Disease}} 
& \multicolumn{1}{c}{\textbf{RHB}} \\ 
\cmidrule(lr){2-5} \cmidrule(lr){6-7} \cmidrule(lr){8-8}
& \multicolumn{2}{c}{\textbf{NTU-60}} 
& \multicolumn{2}{c|}{\textbf{NTU-120}} 
& \textbf{EHE} & \textbf{KIMORE} & \textbf{UI-PRMD} \\
\cmidrule(lr){2-3} \cmidrule(lr){4-5}
& \textbf{\scriptsize{X-Sub}} & \textbf{\scriptsize{X-View}} 
& \textbf{\scriptsize{X-Sub}} & \textbf{\scriptsize{X-Set}} 
&  &  & \\ 
\midrule
R-Ctr   & 93.8 & 95.7 & 91.3 & 86.8 & \textbf{83.15} & \textbf{80.09} & \textbf{87.06} \\
R-Utr  & 94.7 & 95.7 & 91.9 & 86.8 & 81.63 & 77.09 & 87.00 \\
NR-Ctr  & 94.6 & 96.6 & 91.1 & 88.3 & 80.03 & 79.05 & 85.02 \\
NR-Utr & \textbf{95.2} & \textbf{96.8} & \textbf{92.3} & \textbf{88.4} & 79.25 & 77.77 & 84.80 \\
\hline
\end{tabular}
}
\label{transformation_matrices_results}
\end{table}

\noindent\textbf{\uline{Performance Comparison of DML Variants}:} In Tab.~\ref{DML_variants_results_}, we report the performance of each DML variant on the best transformation variant for each domain from Tab.~\ref{transformation_matrices_results}. The results on other transformation variants are provided in Appendix B of the supplementary text. The results in Tab.~\ref{DML_variants_results_} show that all variants improve performance, highlighting the benefit of the proposed DML in reducing distortions. For action datasets, global inhomogeneous is best, reflecting joint-specific variability in skeleton shapes. For disease and rehabilitation datasets, global homogeneous performs best, consistent with their globally constrained and homogeneous movement patterns. 

\begin{table}[th]
\centering
\setlength{\tabcolsep}{1.7pt}
\caption{Performance (\% accuracy) of DML variants on best transformation for each domain from Tab.~\ref{transformation_matrices_results}. Best are bolded. G, L, H and IN are global, local, homogeneous and inhomogeneous.}
{\footnotesize
\begin{tabular}{lcccc|cc|c}
\toprule
\multirow{3}{*}{\textbf{Variants}} 
& \multicolumn{4}{c|}{\textbf{Action}} 
& \multicolumn{2}{c|}{\textbf{Disease}} 
& \multicolumn{1}{c}{\textbf{RHB}} \\ 
\cmidrule(lr){2-5} \cmidrule(lr){6-7} \cmidrule(lr){8-8}
& \multicolumn{2}{c}{\textbf{NTU-60}} 
& \multicolumn{2}{c|}{\textbf{NTU-120}} 
& \textbf{EHE} & \textbf{KIMORE} & \textbf{UI-PRMD} \\
\cmidrule(lr){2-3} \cmidrule(lr){4-5}
& \textbf{\scriptsize{X-Sub}} & \textbf{\scriptsize{X-View}} 
& \textbf{\scriptsize{X-Sub}} & \textbf{\scriptsize{X-Set}} 
&  &  & \\ 
\midrule
GH & 97.0 & 98.5 & 95.0 & 92.7 & \textbf{87.14} & \textbf{85.93} & \textbf{95.19} \\
GIN & \textbf{97.1} & \textbf{98.6} & \textbf{95.2} & \textbf{93.3} & 86.51 & 84.78 & 93.95 \\
LH  & 96.8 & 98.6 & 95.2 & 93.1 & 85.85 & 85.23 & 93.89 \\
LIN & 96.6 & 98.6 & 95.2 & 92.5 & 86.18 & 85.96 & 94.85 \\
\hline
\end{tabular}
}
\label{DML_variants_results_}
\end{table}

\noindent\textbf{\uline{Robustness of DML to different choice of reference shape:}} The results in Tab.~\ref{reference_frame_experiment} show the performance of the proposed DML under different reference frame choices (first, tenth, twentieth, etc) on the NTU-120 X-Sub dataset. The results show that the proposed DML consistently improves performance regardless of the reference frame choice, demonstrating its effectiveness in reducing distortions rather than overfitting to a specific reference pose. For all experiments of this work, we keep the first frame as the choice of reference for consistency.  

\begin{table}[ht]
\centering
\setlength{\tabcolsep}{1.6pt}
\caption{Performance (\% accuracy) of proposed DML on different choice of reference frame on NTU-120 X-Sub dataset.}
{\footnotesize
\begin{tabular}{|c|c|c|c|c|c|c|c|c|c|c|}
\hline
& \textbf{1st} & \textbf{10th} & \textbf{20th} & \textbf{30th}  & \textbf{40th} & \textbf{50th} & \textbf{60th} & \textbf{70th} & \textbf{80th} & \textbf{90th}\\
\hline
without DML & 92.3 & 90.6 & 91.1 & 90.6 & 89.1 & 89.1 & 89.0 & 89.2& 89.0 & 88.8\\
\hline
with DML & 95.2  & 95.0 & 95.3 & 95.0 & 95.0 & 95.2 & 95.1& 95.3& 95.0 & 95.0 \\
\hline
\end{tabular}
}
\label{reference_frame_experiment}
\end{table}

\subsection{Proposed DML versus Parallel Transport (PT)}
\label{comparison_with_pt}

\textbf{\uline{Performance comparison of DML and PT under Choice of Reference as the First Shape}:} We further assess the performance of DML by comparing it with the PT technique (a well-known method for minimizing distortions). For PT, we implement the Pole Ladder (PL) technique from \cite{guigui2022numerical} to transport the tangent representative skeleton between consecutive skeleton shapes to a common reference. Specifically, we first perform log map activation between consecutive shapes in each sequence (that is, we project each current shape to the tangent space of the previous) and then transport these tangent representations to a common reference at the first shape using the PL algorithm. The results of PT are given in the second row of Tab.~\ref{DML_vs_parallel_transp_results}. \\
\indent The first row in Tab.~\ref{DML_vs_parallel_transp_results} corresponds to results under the first shape (FS) as the reference (that is, log-map projection of all skeleton shapes to this same reference) while the third row belongs to adding DML under this same FS. PT demonstrates inconsistent improvements over FS across all datasets, for example, a performance drop behind FS on diseased datasets. DML however consistently improves over FS and shows higher improvements than PT, demonstrating the effectiveness of our learning-based approach. 

\begin{table}[ht]
\centering
\setlength{\tabcolsep}{1.7pt}
\renewcommand{\arraystretch}{0.5}
\caption{Comparison (\% accuracy) of PT and DML under projection at the first shape (FS). Values in blue indicate improvements over FS. * denotes $p \leq 0.05$ between PT and DML.}
{\footnotesize
\begin{tabular}{lcccc|cc|c}
\toprule
\multirow{3}{*}{\textbf{\scriptsize{Protocol}}} 
& \multicolumn{4}{c|}{\textbf{Action}} 
& \multicolumn{2}{c|}{\textbf{Disease}} 
& \multicolumn{1}{c}{\textbf{RHB}} \\ 
\cmidrule(lr){2-5} \cmidrule(lr){6-7} \cmidrule(lr){8-8}
& \multicolumn{2}{c}{\textbf{NTU-60}} 
& \multicolumn{2}{c|}{\textbf{NTU-120}} 
& \textbf{EHE} & \textbf{KIMORE} & \textbf{UI-PRMD} \\
\cmidrule(lr){2-3} \cmidrule(lr){4-5}
& \textbf{\scriptsize{X-Sub}} & \textbf{\scriptsize{X-View}} 
& \textbf{\scriptsize{X-Sub}} & \textbf{\scriptsize{X-Set}} 
&  &  & \\ 
\midrule
FS  & 95.2 & 96.8 & 92.3 & 88.4 & 83.15 & 80.09 & 87.06 \\
\midrule
PT  & \makecell{95.8 \\ \scriptsize $(\uparrow$ \color{blue} 0.9) } & \makecell{98.4 \\ \scriptsize $(\uparrow$ \color{blue} 1.6) } & \makecell{94.3 \\ \scriptsize $(\uparrow$ \color{blue} 2.0)} & \makecell{91.2 \\ \scriptsize $(\uparrow$ \color{blue} 2.8)} & \makecell{78.81 \\ \scriptsize $(\downarrow$ \color{red} 4.34)} & \makecell{73.41 \\ \scriptsize $(\downarrow$ \color{red} 6.68)} & \makecell{89.23 \\ \scriptsize $(\uparrow$ \color{blue} 2.17)} \\
\midrule
DML  & \makecell{\textbf{97.1}\textsuperscript{*} \\ \scriptsize $(\uparrow$ \color{blue} 1.9) } & \makecell{\textbf{98.6}\textsuperscript{*} \\ \scriptsize $(\uparrow$ \color{blue} 1.8)} & \makecell{\textbf{95.2}\textsuperscript{*} \\ \scriptsize $(\uparrow$ \color{blue} 2.9)} & \makecell{\textbf{93.3}\textsuperscript{*} \\ \scriptsize $(\uparrow$ \color{blue} 4.9)} & \makecell{\textbf{87.14}\textsuperscript{*} \\ \scriptsize $(\uparrow$ \color{blue} 3.99)} & \makecell{\textbf{85.93}\textsuperscript{*} \\ \scriptsize $(\uparrow$ \color{blue} 5.84)} & \makecell{\textbf{95.19}\textsuperscript{*} \\ \scriptsize $(\uparrow$ \color{blue} 8.13)} \\
\hline
\end{tabular}
}
\label{DML_vs_parallel_transp_results}
\end{table}

\begin{figure}[th]
    \centering
    \includegraphics[width=\linewidth]{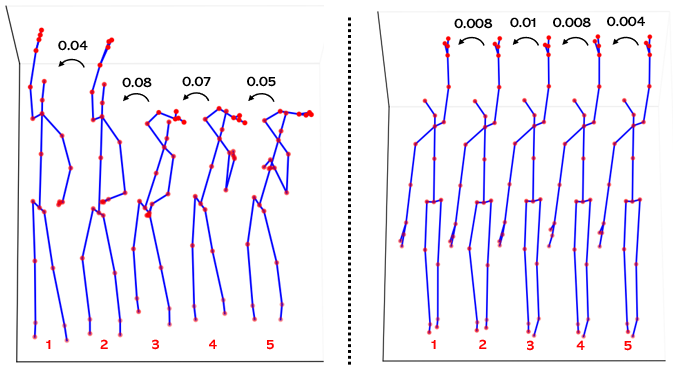}
    \caption{Five consecutive shapes of normal (left) and Alzheimer's subjects (right) for Bend Waist exercise in EHE dataset. Black values are geodesic distance between the current shape and previous.}
    \label{pt_figure}
\end{figure}

\noindent\textbf{\uline{PT Failure on Diseased Datasets}:} Fig. \ref{pt_figure} shows five consecutive shapes for normal and Alzheimer’s subjects performing exercise 4 (bend waist to right) in the EHE dataset, where we observed reduced performance with PT. We remark that contrary to normal subjects, consecutive shapes in Alzheimer’s exhibit minimal displacement (supported by lower geodesic distances), reflecting limited motion amplitude and rigidity in the waist-bending task. Thus, the local tangent representations between consecutive shapes for these subjects are dominated by numerical noise, leading to unstable chained transports and thus hurting performance. 

\subsection{Additional Studies}
\label{additional study}
\noindent\textbf{\uline{Our proposed GTL versus transformation layer (TL) proposed by \cite{friji2021geometric} on NTU-120 Dataset}:} We compare the performance of GTL with the TL introduced in the referenced study under two protocols as shown in Tab.~\ref{transformation_performance_comparison}.  The TL results are taken directly from their ablation studies, and all comparisons are conducted under the same baseline setting for fairness. The first and second rows correspond to TL performance on the non-linear pre-shape space and the linear tangent space, respectively. GTL demonstrates performance gains over TL on both protocols particularly for the non-linear pre-shape setting (35.1\% and 24.6\% gains on X-Sub and X-Set). This confirms the effectiveness of our GTL in accounting for the non-linear underlying geometry of the skeleton sequences in contrast to the TL of \cite{friji2021geometric} that assumes the sequences must reside in a linear space—a limitation that our GTL explicitly removes. 

\begin{table}[ht]
\centering
\setlength{\tabcolsep}{6pt}
\renewcommand{\arraystretch}{0.25}
\caption{Performance comparison of our proposed GTL vs. transformation layer (TL) proposed by \cite{friji2021geometric} on NTU-120 dataset.}
{\footnotesize
\begin{tabular}{l|c|c}
\toprule
\textbf{Protocol} 
& \textbf{\scriptsize{X-Sub (\% acc.) }} & \textbf{\scriptsize{X-Set (\% acc.)}}  \\ 
\midrule
\textbf{\scriptsize{TL by \cite{friji2021geometric} on pre-shape space}} & 57.2 & 63.8 \\
\midrule
\textbf{\scriptsize{TL by \cite{friji2021geometric} on the tangent space}} & 90.6 & 86.7 \\
\midrule
\textbf{\scriptsize{GTL (ours) }} & \makecell{\textbf{92.3}} & \makecell{\textbf{88.4}} \\
\hline
\end{tabular}
}
\label{transformation_performance_comparison}
\end{table}

\begin{figure}[h]
    \centering
    \includegraphics[width=\linewidth]{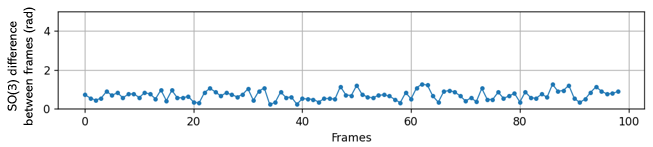}
    \caption{Rotation difference between consecutive frames on NTU-60 X-Sub.}
    \label{SO3_figure}
\end{figure}

\noindent\textbf{\uline{Temporal Coherence Analysis of Skeleton Frames}}: We study the temporal consistency of the rigid-constrained variant of the proposed GTL by computing the differences between optimal rotations for consecutive frames $R_f$ and $R_{f+1}$, computed as $\theta_\text{rad} = \arccos((\mathrm{trace}(R_{f+1} R_f^\top) - 1)/2)$. 
The results, shown in Fig.~\ref{SO3_figure}, indicate that consecutive rotations relax temporal coherence of the skeletons, 
as each rotation independently aligns the frame to the reference. 
This frame-wise alignment finds the geodesic path between each skeleton and the reference, thereby isolating the intrinsic deformation between skeleton shapes. Consequently, while temporal smoothness is not strictly preserved, the computed rotations emphasize shape-specific deformation relevant for motion recognition.

\begin{figure}[th]
    \centering
    \includegraphics[width=\linewidth]{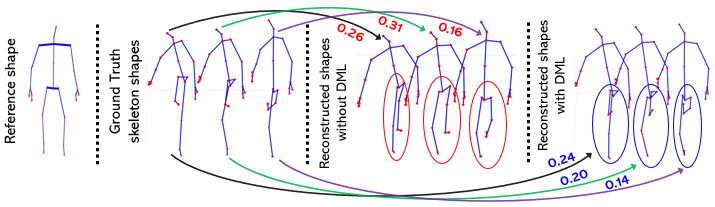}
    \caption{Visualization of DML effect on some representative skeleton shapes for `pick up' action of NTU dataset. Red and blue values are geodesic distances before and after DML respectively.}
    \label{dml_illustra_fig}
\end{figure}

\noindent\textbf{\uline{Visualization of Distortion Reduction by DML:}} Fig. \ref{dml_illustra_fig} illustrates the effect of DML on a test sample of the `pick up’ action from the NTU dataset. Some representative shapes for this action are shown, with a neutral pose used as the reference for projecting these shapes to the tangent space. The red values indicate geodesic distances between the shapes and their tangent-space reconstructions obtained via the exponential map (Eq.~\ref{eq:expmap}), while the blue values correspond to geodesic distances after applying DML. Distortions are observed at the spine base, hips, and left leg (highlighted in red), which are effectively reduced by DML, as evidenced by lower geodesic distances shown in blue.
\section{Conclusion}
\label{conclusion}
In this work, we propose E2E-GNet, a novel end-to-end geometric network for skeleton-based human motion recognition. E2E-GNet integrates a geometric transformation layer that optimizes sequences directly on the non-linear manifold before projecting them to the tangent space, followed by a distortion minimization layer that compensates for projection-induced distortions. Extensive ablations and comparisons validate the effectiveness of both layers, with each variant showing consistent behavior tailored to a respective application. Experiments on five benchmark datasets spanning action recognition, disease analysis, and rehabilitation show that E2E-GNet consistently surpasses state-of-the-art methods. Together, these results highlight the effectiveness, efficiency, and broad applicability of our geometric design in the proposed E2E-GNet.

\section*{Acknowledgments} 
This work of the Interdisciplinary Thematic Institute HealthTech, as part of the ITI 2021-2028 program of the University of Strasbourg, CNRS and Inserm, was supported by IdEx Unistra (ANR-10-IDEX-0002) and SFRI (STRAT’US project, ANR-20-SFRI-0012) under the framework of the French Investments for the Future Program. The authors would like to acknowledge the High Performance Computing Center of the University of Strasbourg for supporting this work by providing scientific support and access to computing resources. Part of the computing resources were funded by the Equipex Equip@Meso project (Programme Investissements d'Avenir) and the CPER Alsacalcul/Big Data.

{
    \small
    \bibliographystyle{ieeenat_fullname}
    \bibliography{main}
}

\setcounter{section}{0}
\renewcommand{\thesection}{\Alph{section}}
\clearpage
\setcounter{page}{1}
\maketitlesupplementary

\section{Inference Time of Proposed Method across Datasets}
In order to support the reported FLOPS and Params of proposed E2E-GNet shown in the main text, we provide in Table~\ref{Inference_time} the full inference time (IT) per sample in each dataset. All reported FLOPS and IT include Maxpool1D, Conv1D, LSTM, fully connected layers, Geometric Transformation layers (GTL), distortion minimization layer (DML), and pre-shape space operations. The results in Table~\ref{Inference_time} indicate a lower inference time which supports the claim that the proposed E2E-GNet is computationally efficient.   
\begin{table}[h]
\centering
\setlength{\tabcolsep}{6pt} 
\renewcommand{\arraystretch}{1.5}
{\footnotesize
\begin{tabular}{|c|c|c|c|c|c|}
\hline
& \textbf{NTU-60/120} & \textbf{EHE} & \textbf{KIMORE} & \textbf{UI-PRMD}\\
\hline
IT (sec.) & 0.004 & 0.003 & 0.0025 & 0.0025 \\
\hline
\end{tabular}
}
\caption{Complete inference time (IT) per sample in each dataset}
\label{Inference_time}
\end{table}

\section{Ablation Results of each DML Variant under each Geometric Transformation Setting}
Table~\ref{DML_variants_results} and~\ref{DML_variants_results_2} present for all datasets, the performance of each distortion minimization layer (DML) variant under each geometric transformation setting. Recall from the main text that the best geometric transformation variant for action recognition datasets is Non-rigid-Unconstrained (NR-Utr), while for disease and rehabilitation datasets is Rigid-Constrained (R-Ctr). Accordingly, in the tables below, the most consistent DML variant under the best transformation setting for each domain is highlighted in bold. 

\begin{table*}[t]
\centering
\setlength{\tabcolsep}{2pt}
{\footnotesize
\begin{tabular}{l|cccc|cccc|cccc|cccc}
\toprule
\multirow{3}{*}{\makecell{\textbf{DML}\\\textbf{Variants}}} 
& \multicolumn{4}{c|}{\textbf{NTU-60 X-Sub}}
& \multicolumn{4}{c|}{\textbf{NTU-60 X-View}}
& \multicolumn{4}{c|}{\textbf{NTU-120 X-Sub}}
& \multicolumn{4}{c}{\textbf{NTU-120 X-Set}} \\
\cmidrule(lr){2-5} \cmidrule(lr){6-9} \cmidrule(lr){10-13} \cmidrule(lr){14-17}
& \textbf{R-Ctr} & \textbf{R-Utr} & \textbf{NR-Ctr} & \textbf{NR-Utr} 
& \textbf{R-Ctr} & \textbf{R-Utr} & \textbf{NR-Ctr} & \textbf{NR-Utr} 
& \textbf{R-Ctr} & \textbf{R-Utr} & \textbf{NR-Ctr} & \textbf{NR-Utr} 
& \textbf{R-Ctr} & \textbf{R-Utr} & \textbf{NR-Ctr} & \textbf{NR-Utr}  \\
\midrule
GH   & 96.2 & 96.5 & 96.4 & 97.0 & 98.4 & 98.7 & 98.3 & 98.5 & 93.9 & 94.6 & 95.0 & 95.0 & 91.1 & 91.9 & 92.3 & 92.7 \\
GIN  & 97.0 & 97.0 & 97.0 & \textbf{97.1} & 98.5 & 98.8 & 98.4 & \textbf{98.6} & 95.0 & 95.4 & 95.1 & \textbf{95.2} & 92.0 & 92.9 & 93.0 & 
\textbf{93.3} \\
LH & 96.5 & 96.6 & 96.6 & 96.8 & 98.5 & 98.6 & 98.3 & 98.6 & 95.1 & 94.9 & 95.0 & 95.2 & 91.3 & 92.0 & 93.0 & 93.1 \\
LIN  & 96.1 & 96.5 & 96.1 & 96.6 & 98.2 & 98.5 & 98.7 & 98.6 & 94.5 & 95.1 & 94.9 & 95.2 & 92.2 & 92.1 & 92.3 & 92.5 \\
\hline
\end{tabular}
}
\caption{Performance (\% accuracy) of each DML variant under each geometric transformation setting for action recognition datasets. G, L, H and IN are global, local, homogeneous and inhomogeneous respectively. R-Ctr, R-Utr, NR-Ctr and NR-Utr denote rigid-constrained, rigid-unconstrained, non-rigid-constrained and non-rigid-unconstrained respectively.}
\label{DML_variants_results}
\end{table*}
\begin{table*}[t]
\centering
\setlength{\tabcolsep}{6pt}
{\footnotesize
\begin{tabular}{l|cccc|cccc|cccc}
\toprule
\multirow{3}{*}{\makecell{\textbf{DML}\\\textbf{Variants}}} 
& \multicolumn{4}{c|}{\textbf{EHE}}
& \multicolumn{4}{c|}{\textbf{KIMORE}}
& \multicolumn{4}{c}{\textbf{UI-PRMD}} \\
\cmidrule(lr){2-5} \cmidrule(lr){6-9} \cmidrule(lr){10-13}
& \textbf{R-Ctr} & \textbf{R-Utr} & \textbf{NR-Ctr} & \textbf{NR-Utr} 
& \textbf{R-Ctr} & \textbf{R-Utr} & \textbf{NR-Ctr} & \textbf{NR-Utr} 
& \textbf{R-Ctr} & \textbf{R-Utr} & \textbf{NR-Ctr} & \textbf{NR-Utr}   \\
\midrule
GH  & \textbf{87.14} & 86.74 & 86.62 & 87.00 & \textbf{85.93} & 85.19 & 84.81 & 85.31 & \textbf{95.19} &  92.60 & 92.93 & 93.95 \\
GIN &86.51 & 86.27 & 85.24 & 85.77 & 84.78 & 84.49 & 83.78 & 84.81 & 93.95  & 91.47 & 91.38 & 95.47 \\
LH &85.85 & 86.62 & 85.55 & 86.63 & 85.23 & 85.09 & 84.88 & 85.84 &  93.89 & 92.64 & 92.63 & 95.61 \\
LIN & 86.18 & 85.46 & 85.90 & 85.86 & 85.96 & 83.75 & 83.67 & 83.16 &  94.85 & 92.54 & 93.39 & 95.76\\
\hline
\end{tabular}
}
\caption{Performance (\% accuracy) of each DML variant under each geometric transformation setting for disease and rehabilitation datasets. G, L, H and IN are global, local, homogeneous and inhomogeneous respectively. R-Ctr, R-Utr, NR-Ctr and NR-Utr denote rigid-constrained, rigid-unconstrained, non-rigid-constrained and non-rigid-unconstrained respectively.}
\label{DML_variants_results_2}
\end{table*}

\section{Detailed Cross Validation Folds Division}
In Table \ref{ehe_kimore_ui_prmd_folds} below, we provide for the three datasets (EHE, KIMORE and UI-PRMD) the number of available exercise sequences present in the testing set of each cross-validation fold. The same cross-validation split is used by all other compared methods in this work.
\begin{table*}[!t]
\centering
\setlength{\tabcolsep}{4.3pt} 
\renewcommand{\arraystretch}{2.5}
{\footnotesize
\begin{tabular}{|c|c|c|c|c|c|c|c|c|c|c|c|c|c|c|c|c|c|c|c|c|c|c|c|c|c|c|}
\hline
\makecell{\textbf{Folds}} & \multicolumn{6}{c|}{\textbf{EHE}} & \multicolumn{8}{c|}{\textbf{KIMORE}} & \multicolumn{10}{c|}{\textbf{UI-PRMD}}\\
\hline
& \textbf{E1} & \textbf{E2} & \textbf{E3} & \textbf{E4} & \textbf{E5} & \textbf{E6} 
& \textbf{Es1} & \makecell{\textbf{Es2} \\ \textbf{(L)}} & \makecell{\textbf{Es2} \\ \textbf{(R)}} & \makecell{\textbf{Es3} \\ \textbf{(L)}} & \makecell{\textbf{Es3} \\ \textbf{(R)}} & \makecell{\textbf{Es4} \\ \textbf{(L)}} & \makecell{\textbf{Es4} \\ \textbf{(R)}} & \textbf{Es5} 
& \textbf{E1} & \textbf{E2} & \textbf{E3} & \textbf{E4} & \textbf{E5} & \textbf{E6} & \textbf{E7} & \textbf{E8} & \textbf{E9} & \textbf{E10} \\
\hline
\textbf{F1} & 33 & 27 & 27 & 30 & 14 & 14 & 58 & 53 & 54 & 54 & 51 & 47 & 56 & 58 & 38 & 30 & 30 & 18 & 36 & 20 & 18 & 16 & 20 & 36 \\
\hline 
\textbf{F2} & 34 & 27 & 35 & 26 & 12 & 12 & 55 & 47 & 44 & 48 & 48 & 38 & 61 & 51 & 38 & 34 & 18 & 36 & 34 & 38 & 18 & 28 & 18 & 18 \\
\hline
\textbf{F3} & 41 & 30 & 42 & 35 & 16 & 16 & 49 & 34 & 34 & 38 & 39 & 49 & 34 & 53 & 34 & 14 & 18 & 18 & 36 & 18 & 18 & 16 & 18 & 18 \\
\hline
\textbf{F4} & 45 & 30 & 52 & 52 & 16 & 15 & 43 & 31 & 34 & 37 & 38 & 33 & 43 & 48 & 36 & 14 & 18 & 36 & 34 & 32 & 36 & 30 & 36 & 18 \\
\hline
\textbf{F5} & 45 & 30 & 40 & 43 & 16 & 14 & 50 & 31 & 35 & 33 & 33 & 56 & 26 & 45 & 34 & 18 & 18 & 32 & 28 & 38 & 36 & 36 & 28 & 18 \\
\hline
\end{tabular}
}
\caption{Exercise Repetitions in the test set of each fold for EHE, KIMORE and UI-PRMD datasets. E and Es represent Exercise while R and L are right and left respectively.}
\label{ehe_kimore_ui_prmd_folds}
\end{table*}

\section{Additional Evaluation Using Model's Predicted Probability Scores on EHE and UI-PRMD Datasets}
Alongside classification accuracy, separation degree (SD), distance metrics (DM), cross correlation (CR), and Euclidean distance (ED) have been employed in previous works to assess a model's prediction ability and to quantify the consistency between machine and human evaluation. \\ 
\textbf{Model's Prediction Capacity:} Studies in \cite{bruce2024egcn++, liao2020deep, williams2019assessment, yu2022egcn} examined a model's representation capacity with SD and DM. SD and DM are used to quantify the difference between correct and incorrect predicted evaluation scores. Similar to these studies, we compute SD and DM for the UI-PMRD dataset where expert evaluation scores are not available. Given a pair of positive numbers $x$ and $y$, SD is computed as $S_{D}(x,y) = \frac{x-y}{x+y} \in [-1,1]$. For two positive sequences $x = (x_{1}, ..., x_{y})$ and $y = (y_{1}, ..., y_{z})$, SD and DM are calculated as:
        \begin{equation}
             S_{D}(x,y) = \frac{1}{yz}\sum_{i=1}^{y}\sum_{j=1}^{z}S_{D}(x_{i},y_{j})
        \end{equation}
 
        \begin{equation}
             D_{M}(x_{n},y_{n}) = \frac{|x_{n} - y_{n}|}{\sqrt{\dfrac{1}{y}\sum_{i=1}^{y}(x_{n} -y_{n})^2}}
        \end{equation}

\noindent \textbf{Consistency between Human and Model's Evaluation:} CR and ED have been employed in \cite{bruce2024egcn++, bruce2021skeleton} to study the consistency between expert and model's evaluation scores. CR and ED quantify the differences between model's predicted numerical scores and expert's clinical scores. Lower ED and higher CR tell that a model's evaluation score is more consistent with human's and vice versa. Given two vectors $x = (x_{1}, x_{2}, ..., x_{j})$ and $y = (y_{1}, y_{2}, ..., y_{j})$, the CR and ED between them are computed as:
        \begin{equation}
             C_{R}(x,y) = \frac{\sum_{i=1}^{j}(x-\bar{x})(y-\bar{y})}{\sqrt{\sum_{i=1}^{j}(x-\bar{x})^2 \sum_{i=1}^{j}(y-\bar{y})^2}}
        \end{equation}
 
        \begin{equation}
             E_{D}(x,y) = \sqrt{\sum\nolimits_{i=1}^{j}(x_{i} - y_{i})^2}
        \end{equation}

\noindent For these calculations, we took the probability scores before the final classification layer as similarly done in all compared methods. These scores can serve as model's numerical evaluation scores and we evaluate their consistency with human's evaluation scores on the EHE dataset where expert's screening scores are available. The EHE dataset \cite{bruce2021skeleton} provides clinical scores ranging from 0 to 10 indicating varying severity of Alzheimer's disease. However, for the UI-PRMD dataset where no clinical scores are present, we evaluate the model's confidence in separating between the correct and incorrect predictions. The results of these evaluations are given in Table~\ref{all_evaluation_metrics_table}.

\begin{table*}[t]
\centering
\setlength{\tabcolsep}{10.0pt} 
\renewcommand{\arraystretch}{1.7}
{\footnotesize
\begin{tabular}{|c|c|c|c|c|c|c|c|c|}
\hline
\textbf{Methods} & \multicolumn{4}{c|}{\textbf{EHE}} & \multicolumn{4}{c|}{\textbf{UI-PRMD}} \\
\hline
\textbf{} & \multicolumn{2}{c|}{\textbf{Sigmoid}} & \multicolumn{2}{c|}{\textbf{Softmax}} & \multicolumn{2}{c|}{\textbf{Sigmoid}} & \multicolumn{2}{c|}{\textbf{Softmax}} \\
\hline
& $\mathbf{E_{D}}$ $\Downarrow$ & $\mathbf{C_{R}}$ $\Uparrow$ & $\mathbf{E_{D}}$ $\Downarrow$ & $\mathbf{C_{R}}$ $\Uparrow$ & $\mathbf{S_{D}}$ $\Uparrow$ & $\mathbf{D_{M}}$ $\Uparrow$ & $\mathbf{S_{D}}$ $\Uparrow$ & $\mathbf{D_{M}}$ $\Uparrow$ \\
\hline
\textbf{GCN} \cite{bruce2020skeleton}, \cite{bruce2021skeleton} & 1.7357 & 0.3458 & 1.5382 & 0.4878 & 0.3035 & 0.7736 & 0.4336 & 0.7830 \\
\hline
\textbf{KShapeNet} \cite{friji2021geometric} & 1.7697 & \uline{0.6732} & 1.7856 & 0.6625 & \uline{0.5772} & \uline{0.9284} & 0.4626 & 0.8895 \\
\hline
\textbf{EGCN} \cite{yu2022egcn} & \uline{1.5893} & 0.5111 & \uline{1.5077} & 0.5531 & 0.4142 & 0.8280 & 0.5307 & 0.8494 \\
\hline
\textbf{EGCN++} \cite{bruce2024egcn++} & \textbf{1.5456} & 0.6254 & \textbf{1.4444} & \uline{0.6713} & 0.5419 & 0.8702 & \uline{0.6485} & \uline{0.8916} \\
\hline
\makecell{\textbf{E2E-GNet}} & \makecell{1.6312} & \makecell{\textbf{0.7251}} & \makecell{1.5952} & \makecell{\textbf{0.7384}} & \makecell{\textbf{0.7447}} & \makecell{\textbf{0.9587}} & \makecell{\textbf{0.7285}} & \makecell{\textbf{0.9424}} \\
\hline
\end{tabular}
}
\caption{$\mathbf{E_{D}}$, $\mathbf{C_{R}}$, $\mathbf{S_{D}}$, and $\mathbf{D_{M}}$ results for EHE and UI-PRMD datasets, with comparison to other state-of-the-art methods. Best results are in bold, second best underlined.}
\label{all_evaluation_metrics_table}
\end{table*}

\section{Classification Results Per Exercise on EHE, KIMORE and UI-PRMD Datasets}
Table~\ref{EHE_per_exercise}, \ref{KIMORE_per_exercise}, and \ref{UI-PRMD_per_exercise} provides respectively the classification accuracy (per exercise) of our proposed E2E-GNet for EHE, KIMORE and UI-PRMD Datasets. 

\begin{table*}[th]
\centering
\setlength{\tabcolsep}{7.0pt}
\renewcommand{\arraystretch}{1.5}
{\footnotesize
\begin{tabular}{|c|c|c|c|c|c|c|c|}
\hline
\makecell{\textbf{Dataset}} & \multicolumn{6}{c|}{\textbf{Exercise ID}} & \textbf{Average} \\
\hline
& \makecell{\textbf{E1}} 
& \makecell{\textbf{E2}} 
& \makecell{\textbf{E3}} 
& \makecell{\textbf{E4}}
& \makecell{\textbf{E5}}
& \makecell{\textbf{E6}} 
& {}\\
\hline
\makecell{\textbf{EHE}} & 89.51  & 88.52 & 88.63 & 83.60 & 90.36 & 82.24 & \textbf{87.14} \\
\hline 
\end{tabular}
}
\caption{Performance results (\% accuracy per exercise) of the proposed E2E-GNet for the EHE Dataset.}
\label{EHE_per_exercise}
\end{table*}

\begin{table*}[th]
\centering
\setlength{\tabcolsep}{7.0pt} 
\renewcommand{\arraystretch}{1.5} 
{\footnotesize
\begin{tabular}{|c|c|c|c|c|c|c|c|c|c|}
\hline
\makecell{\textbf{Dataset}} & \multicolumn{8}{c|}{\textbf{Exercise ID}} & \textbf{Average} \\
\hline
& \makecell{\textbf{Es1}} 
& \makecell{\textbf{Es2(L)}} 
& \makecell{\textbf{Es2(R)}} 
& \makecell{\textbf{Es3(L)}}
& \makecell{\textbf{Es3(R)}}
& \makecell{\textbf{Es4(L)}} 
& \makecell{\textbf{Es4(R)}} 
& \makecell{\textbf{Es5}} 
& {}\\
\hline
\makecell{\textbf{KIMORE}} & 86.97  & 81.61 & 88.06 & 87.38 & 90.33 & 84.79 & 84.11 & 84.21 & \textbf{85.93} \\
\hline 
\end{tabular}
}
\caption{Performance results (\% accuracy per exercise) of the proposed E2E-GNet for the KIMORE Dataset.}
\label{KIMORE_per_exercise}
\end{table*}

\begin{table*}[th]
\centering
\setlength{\tabcolsep}{7.0pt}
\renewcommand{\arraystretch}{1.5}
{\footnotesize
\begin{tabular}{|c|c|c|c|c|c|c|c|c|c|c|c|}
\hline
\makecell{\textbf{Dataset}} & \multicolumn{10}{c|}{\textbf{Exercise ID}} & \textbf{Average} \\
\hline
& \makecell{\textbf{E1}} 
& \makecell{\textbf{E2}} 
& \makecell{\textbf{E3}} 
& \makecell{\textbf{E4}}
& \makecell{\textbf{E5}}
& \makecell{\textbf{E6}} 
& \makecell{\textbf{E7}} 
& \makecell{\textbf{E8}} 
& \makecell{\textbf{E9}} 
& \makecell{\textbf{E10}} 
& {}\\
\hline
\makecell{\textbf{UI-PRMD}} & 93.94  & 95.88 & 94.56 & 100 & 100 & 100 & 93.33 & 93.18 & 90.56 & 90.40 & \textbf{95.19} \\
\hline 
\end{tabular}
}
\caption{Performance results (\% accuracy per exercise) of the proposed E2E-GNet for the UI-PRMD Dataset.}
\label{UI-PRMD_per_exercise}
\end{table*}

\end{document}